# Examining stability of machine learning methods for predicting dementia at early phases of the disease

Sinan Faouri[a], Mahmood AlBashayreh[b*] and Mohammad Azzeh[c]

[a]Department of Mechanical and Industrial Engineering, Applied Science Private University, Jordan
[b]Department of Computer Science, Applied Science Private University, Jordan
[c]Department of Data Science, Princess Sumaya University for Technology, Jordan

| C H R O N I C L E | A B S T R A C T |
|---|---|
|  | Dementia is a neuropsychiatric brain disorder that usually occurs when one or more brain cells stop working partially or at all. Diagnosis of this disorder in the early phases of the disease is a vital task to rescue patients' lives from bad consequences and provide them with better healthcare. Machine learning methods have been proven to be accurate in predicting dementia in the early phases of the disease. The prediction of dementia depends heavily on the type of collected data which usually are gathered from Normalized Whole Brain Volume (nWBV) and Atlas Scaling Factor (ASF) which are normally measured and corrected from Magnetic Resonance Imaging (MRIs). Other biological features such as age and gender can also help in the diagnosis of dementia. Although many studies use machine learning for predicting dementia, we could not reach a conclusion on the stability of these methods for which one is more accurate under different experimental conditions. Therefore, this paper investigates the conclusion stability regarding the performance of machine learning algorithms for dementia prediction. To accomplish this, a large number of experiments were run using 7 machine learning algorithms and two feature reduction algorithms namely, Information Gain (IG) and Principal Component Analysis (PCA). To examine the stability of these algorithms, thresholds of feature selection were changed for the IG from 20% to 100% and the PCA dimension from 2 to 8. This has resulted in $7 \times 9 + 7 \times 7 = 112$ experiments. In each experiment, various classification evaluation data were recorded. The obtained results show that among seven algorithms the support vector machine and Naïve Bayes are the most stable algorithms while changing the selection threshold. Also, it was found that using IG would seem more efficient than using PCA for predicting Dementia. These promising results open the door to a new era of early prognosis of Alzheimer's Disease and Related Dementias (ADRD). |

## 1. Introduction

Dementia is a neuropsychiatric brain disorder that usually occurs when one or more brain cells stop working partially or at all. This brain disorder is often accompanied by memory attenuation. From healthcare records, it was found that people aged 65 and above are more vulnerable to disease (Bansal et al., 2018, 2020; Lakshmi, 2020). According to the World Health Organization (WHO), there are around 50 million patients with dementia worldwide with an increase of 10 million patients annually (Battineni et al., 2020; Harvey et al., 2003; WHO, 2020). The most common type of dementia is Alzheimer's Disease (AD) where the average age of clinical AD diagnosis received is 80 years old (Barnes et al., 2015; Dai et al., 2020). The prognosis of dementia is still poor and depends on various factors such as age, gender, educational level, and many more (van de Vorst et al., 2020). Early studies suggested that early diagnosis of dementia may prevent deterioration and help widely with treatments and future predictions of the disease (Alam et al., 2016; Battineni et al., 2019; Chen & Herskovits, 2010). This is where advanced computational techniques come into use for predicting future outcomes of dementia whether as a prognosis of disease progression or as a mortality rate (Green & Zhang, 2016; van de Vorst et al., 2020). At the beginning and mild stages of dementia, magnetic resonance imaging (MRI), which is a neuroimaging technique, is becoming an effective tool in detecting AD. However, only a few works correlated AD incidence rate with measurements concluded from MRI (Battineni et al., 2020). It's worthwhile to note that MRI provides significant input variables for machine learning algorithms to make the prediction and classification of probable dementia patients and all age-related cognitive decline (ARCD) in general (Babiloni et al., 2017; Garrard et al., 2014; Pellegrini et al., 2018). Estimated total intracranial volume (eTIV) is a significant feature in the dataset that was studied in this research, which relates the brain and intracranial size to ADRD in a volumetric method which is common in neurodegenerative diseases analysis (Malone et al., 2014). FAST[1] program of FSL software suite was used for computation of normalized whole brain volume (nWBV) expressed as a percentage of the accumulated voxels of white and grey matter in the brain taken from eTIV analysis where volumetric measures are normalized according to head's size (Battineni et al., 2020; Marcus et al., 2010). Machine learning methods are becoming a support decision tool that can help doctors in diagnosing dementia-related diseases. The dementia prediction is treated as a classification problem where possible output labels are {demented or non-demented}. In the literature, it was found that machine learning algorithms were used in many studies to predict the presence of dementia in diagnosed patients. The conclusion drawn from these studies is confusing regarding the performance of machine learning methods. Therefore, this study examines the stability of seven common machine algorithms for dementia prediction. These algorithms are AdaBoost (Ada) and Random Forest (RF), *k*-Nearest Neighbor (kNN), Support Vector Machine (SVM), Decision Tree (DT), Naïve Bayes (NB), and Logistic Regression (LR). To accomplish the stability examination, two feature reduction algorithms were used: Information Gain (IG) and Principal Component analysis. Information Gain can help in ranking features according to IG score then a specified threshold can be applied to select top-ranked features (KENT, 1983; Raileanu & Stoffel, 2004). Whereas, PCA is a feature reduction algorithm that aims at generating new dimensionality (usually less than the original dimension) based on the structure of the original dataset (Abdi & Williams, 2010; Pearson, 1901; Ringnér, 2008). For each feature reduction algorithm, the threshold of features was changed to produce a new feature subset every time.

Regarding IG, the Information gain score is first measured between each input feature and the output feature. The input features are then ranked according to their Information gain score from high to low. Later, specific thresholds (range from 20% to 100%) are applied to select the top features that will be used to slice the dataset that will be used to train and test machine learning methods. For example, the threshold of 20% means selecting the top 20% of the dataset's features. Every time the threshold was changed, a new set of features was added to the previously selected top features. This will form new input data to the machine learning models which enable us to examine the stability of these models based on different data. On the other hand, the same scenario was used with PCA, where a dimension reduction threshold was specified from 2 dimensions to 8.

A large number of experiments have been conducted to examine the stability of seven machine learning methods in predicting dementia. The stability term is defined by how much the performance of the machine learning method can remain stable under changing different setting parameters. The setting parameters in this study are the changes in the dataset features and dimensions as explained earlier. However, in each experiment, seven machine learning methods were run over a different feature set either obtained from IG or PCA. This has resulted in $7\times9 + 7\times7 = 112$ experiments. The classification accuracy values have been recorded for each experiment. The dementia dataset the has been used in this paper is available publicly from Boysen (2017), more details can be found in section 3.1. The obtained results show that among seven algorithms the support vector machine and Naïve Bayes are the most stable algorithms while changing the selection threshold. Also, the researchers found that using IG would seem more efficient than using PCA for predicting Dementia.

The remainder of this paper is organized as follows. In section 1 an introduction to dementia and machine learning integration is followed by the literature review and related studies in section 2. Section 3 covers experimental setup through four subsections: data origin and dictionary, data preprocessing, choice of learning algorithms, and evaluation measures. In section 4, the research methodology was explained. Moreover, section 5 consists of results and discussion, while section 6 wraps up this study with a conclusion.

## 2. Literature review

Most of the studies performed on the prediction of dementia using machine learning models did not mention a clear methodology that could be generalized to ensure a stable high-performance prediction independently of the underlying dataset (Bansal et al., 2020; Battineni et al., 2020; Bharanidharan & Rajaguru, 2020; Dallora et al., 2020; Popuri et al., 2020; Sharma et al., 2020; You et al., 2019). This is because each study uses a different dataset with various features from different sources. In other words, there is a lack of stability assessment of the proposed models for predicting dementia. In Battineni et al. (2020), a hybrid machine learning model combining 4 models with 14 features to diagnose the early stages of Alzheimer's disease was proposed. The researchers studied 373 magnetic resonance imaging tests belonging to 150 elders. As a result, it was found that the proposed hybrid model provides enhanced accuracy of dementia prediction up to 98%. In another study, a machine learning model was presented for the classification of dementia disease using magnetic resonance imaging (Bansal et al., 2020). The researchers proposed using the bag of features method to extract the features of magnetic resonance imaging scans in association with a support vector machine to classify these scans. This proposed methodology leads to an accuracy of 93% for the detection of dementia. A broad multifactorial decision tree model for the prediction of dementia was proposed by Dallora et al. (2020). The researchers used longitudinal data incorporating 75 variables from 726 subjects. The proposed approach reached 74.5% Area Under the Curve (AUC) for the 10-year prognosis of dementia, while the Recall value, that is the ratio of correctly predicted positive observations, was 72.2%. In the study of Sharma et al. (2020), dementia was diagnosed by implementing an iterative filtering decomposition approach to improve the classification accuracy of electroencephalogram (EEG) signal besides Cognitive Tests including finger tapping test (FTT) and the continuous performance test (CPT). The EEG data were collected from 47 subjects. The proposed approach achieved up to 92% accuracies for dementia classification, 91.67% accuracies for early dementia classification and 91.87% accuracies for healthy classifications. Four swarm intelligence algorithms (particle swarm optimization, artificial bee colony, ant colony optimization, and dragonfly algorithm) were implemented and compared with the non-swarm intelligence algorithm (fuzzy-C-means) as in the study of Bharanidharan and Rajaguru (2020). The brain Cross-sectional MRI of 65 non-dementia and 52 dementia subjects were collected and used as input for dementia classification. The results show that the Dragonfly-particle swarm optimization hybrid classifier yields the highest accuracy of 87.18%. The researchers You et al. (2019), examined the relationship between speech and the risk of dementia using a parallel classification system. The system ensembles both the K-Nearest-Neighbor model and the Support Vector Machine model to classify older participants at high or low risk of dementia by selecting 27-feature extracted from audio recordings. The resulting accuracy was 94.7% when the system trained with paralinguistic features only and 97.2% when the system trained with both paralinguistic and episodic memory test features. The authors Popuri et al. (2020), proposed an ensemble-learning model that combines structural features to create an aggregate measure of neurodegeneration in the brain. The classifier was trained on 753 subjects, including 423 stable normal control and 330 dementia of Alzheimer's type. The classifier was trained on 753 subjects, including 423 stable normal control and 330 dementia of Alzheimer's type. Then, independent validation was made on 8834 unseen images to predict the development of dementia of Alzheimer's type depending on the time-to-conversion. The classification performance achieved an area-under-curve (AUC) of 81% for time-to-conversion of 6 months and 73% for time-to-conversion of 7 years.

---

[1] FAST is a software used for computation of normalized whole brain volume.

## 3. Experimental Setup

This section describes the research methodology of the present paper and is divided into the following subsections:

### 3.1 Data origin and dictionary

The original data came from the OASIS project which aims to provide the scientific community with open-source MRI datasets for free. OASIS is made available by the Washington University Alzheimer's Disease Research Center, Dr. Randy Buckner at the Howard Hughes Medical Institute (HHMI) at Harvard University, the Neuroinformatics Research Group (NRG) at Washington University School of Medicine, and the Biomedical Informatics Research Network (BIRN). However, we've downloaded it in a formatted way from Kaggle (Boysen, 2017). This dataset is based on a study of 150 subjects with multiple visits and scans aged between 60-96. All study participants are right-handed. According to this dataset, subjects can be initially divided into three groups: The first group includes 72 subjects who were classified as non-demented throughout the study. The second group includes 14 subjects who were first diagnosed as non-demented and then converted at a later stage into dementia. The third and final group consists of 64 of the subjects characterized as demented from early visits and remained so for the rest of the study. The last two groups were merged into one group to avoid duplicate results as will be explained in the next section. People with normal age-related brain changes such as mild atrophy, Leukoaraiosis, and common dementia cases of AD, were not excluded from this study. Timeframe for all MRI sessions was within one year. The data dictionary and description for every feature are detailed in Table 1.

**Table 1**
Data dictionary of 15 features and 1 target

| Feature | Description |
|---|---|
| Subject ID | Subject identification |
| MRI ID | MRI exam identification |
| Group | Class (Has disease or not) |
| Visit | Visit order (How many times the patient visited the clinic) |
| MR delay | MR delay time (contrast) |
| M/F | Gender |
| Hand | All participants in the study were right-handed |
| Age | Age |
| EDUC | Years of education |
| SES | Socio-economic status (the social standing or class of an individual or group. It is often measured as a combination of education, income, and occupation. Examinations of socioeconomic status often reveal inequities in access to resources, plus issues related to privilege, power, and control). (As assessed by the Hollingshead Index of Social Position and classified from 1 (highest status) to 5 (lowest status)) |
| MMSE | Mini-Mental State Examination score (range is from 0 = worst to 30 = best) |
| CDR | Clinical Dementia Rating (0 = no dementia, 0.5 = very mild AD, 1 = mild AD, 2 = moderate AD) |
| eTIV | Estimated total intracranial volume, $mm^3$ |
| nWBV | Normalized whole-brain volume, expressed as a percent of all voxels in the atlas-masked image that are labeled as gray or white matter by the automated tissue segmentation process |
| ASF | Atlas scaling factor (unitless). A computed scaling factor that transforms the native-space brain and skull to the atlas target (i.e., the determinant of the transform matrix). |

336

Fig. 1 shows the distribution of the target variable which has two labels {demented, non-demented} grouped by gender. It's clear in Figure 1 that there is a class balance scenario in this binary classification situation. However, it's worth noting that the "converted" label from the dataset was appended and modified to be "demented" because not doing so means having duplicate examples classified as "non-demented" first, then changed to "converted" at a later visit. The latter observation has been noted cautiously and found to have a high impact on the machine learning algorithm performance in subsequent sections in this work. The feature titled "number of visits" was kept in the dataset because of its importance in completing the big picture of the diagnosis procedure. Finally, one can take a baseline idea that males are more exposed to dementia than females according to the studied dataset.

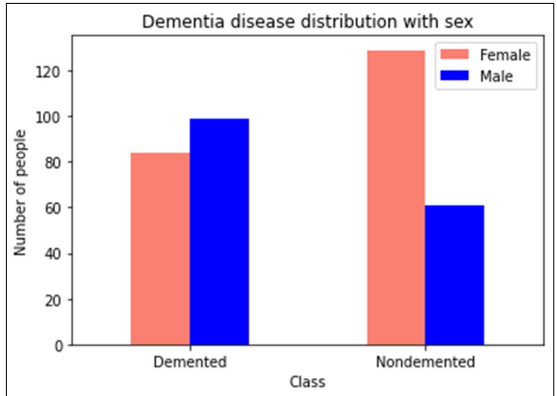 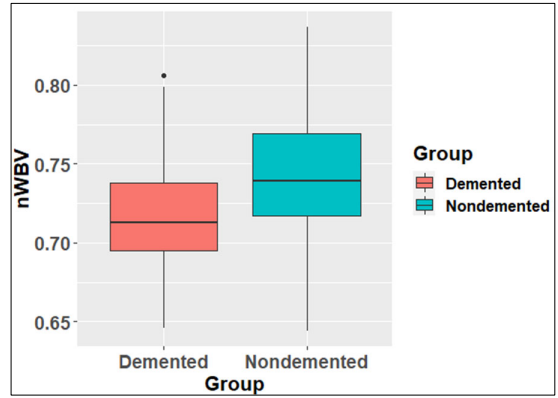

**Fig. 1.** Dementia disease distribution with gender among study participants

**Fig. 2.** Dementia disease distribution with gender among study participants

Fig. 2 shows boxplots of the normalized Whole Brain Volume (nWBV) feature concerning two target labels. It can be observed that the median of nWBV for demented patients is smaller than non-demented patients. Two groups are significantly different, the p-value=0.003. Correlation between age and normalize whole brain volume (nWBV) is shown in Fig. 3. There is a negative correlation between nWBV and age which confirms that the whole brain volume is decreasing as long as the age of patients is increasing.

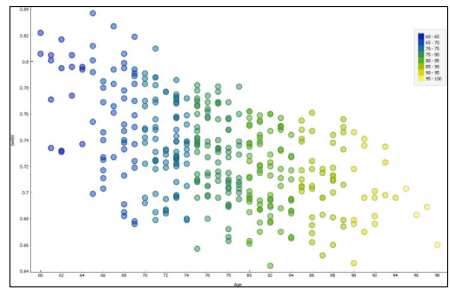

**Fig. 3.** Correlation between age and normalized whole brain volume (nWBV)

Fig. 4 shows the age histogram distribution for all records in the dataset in addition to boxplots of age features concerning target labels. The sample of patients in this dataset is normally distributed where most of the patients aged between 70 and 85. Most of the demented patients aged between 75 and 85 whereas non-demented patients aged between 72 and 82. However, no significant difference between the two groups in terms of age was found. Therefore, one could not conclude that there is a clear relationship between age and dementia, even though the medical sectors proved that there is strong evidence that there is a relationship between age and dementia.

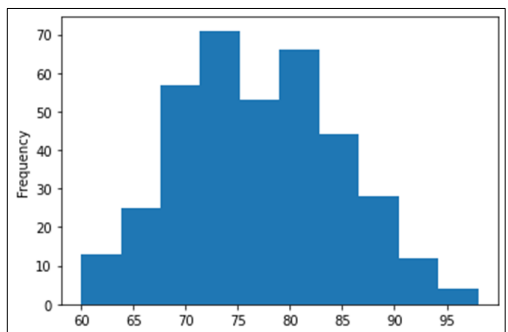 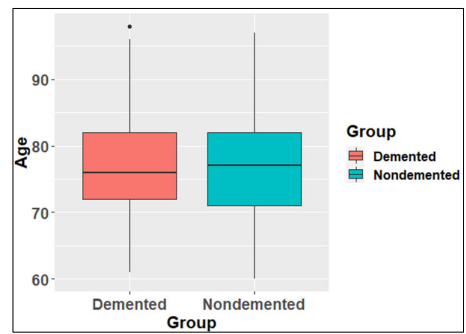

(a) Histogram of Age  (b) Boxplots of Age concerning Target

**Fig. 4.** Age Distribution

*3.2 Data preprocessing*

Data preprocessing is an essential step in constructing any machine learning model to ensure the quality of the training data. In this study, some important data preprocessing steps were carried out such as handling missing values, outlier detection, and data normalization. Missing data in medical records is not unusual (Zriqat et al., 2017). To handle missing values, the mean imputation method was applied for every missing field in the dataset. In other words, the missing values in any feature are replaced with the mean of that feature. Even though there are different other imputation methods, the researchers prefer to use this approach due to its simplicity, low processing cost, and performance in boosting machine learning models. On the other hand, since the numeric features in the dataset present different scales, they may have a negative impact on the performance of the constructed machine learning models. To eliminate such impact from different feature types, and to have the same influence degree the min-max normalization was used. Regarding outliers, no extreme values were found in all features, therefore no observations were excluded.

*3.3 Choice of learning algorithms*

As mentioned in the introduction, various machine learning algorithms were chosen in this study to build different prediction models for predicting the Dementia group. These methods are supposed to exhibit different prediction mechanisms and support both linear and nonlinear relationships between the output variable and all input descriptive variables. In particular, seven common machine learning algorithms were used, some of them are known as ensemble learning algorithms such as AdaBoost (Ada) and Random Forest (RF), while the remaining algorithms are considered solo algorithms such as *k*-Nearest Neighbor (kNN), Support Vector Machine (SVM), Decision Tree (DT), Naïve Bayes (NB), and Logistic Regression (LR). Ensemble learning algorithms are supposed to have better accuracy in comparison with solo algorithms as confirmed in previous studies because they implemented multiple solo algorithms to support weak learners (Minku et al., 2013).

*k*-Nearest Neighbor (kNN) is a supervised machine learning algorithm based on the idea of predicting by similarity. Specifically, the kNN uses a distance measure such as Euclidean distance to retrieve, for each new observation, the nearest observations from the training dataset. The final output is then predicted from the outputs of the set of selected observations. Many control parameters impact the performance of kNN such as feature selection, voting mechanism, feature weighting, number of selected observations ($k$), and type of distance measure. It is important to note that a smaller value of $k$ can be noisy and subject to the effect of an outlier, whereas the large value of $k$ can smooth the final decision (Wu et al., 2002).

SVM is an algorithm used for regression and classification. The basic idea of SVM is to build an optimal hyperplane that can separate data with maximum margin. The margin is defined as the maximal width of the slab parallel to the hyperplane that has no interior data points. The optimal hyperplane generation depends on the choice of kernel functions such as Gaussian, Polynomial, and Radial Basis Function. Both Gaussian and Radial Basis function kernels can benefit hyperplane generation because they support the locality of training data, which means that the data can be efficiently separated (Widodo & Yang, 2007). DT is a tree-like method that uses iterative partitioning to construct the tree. The algorithm uses probabilistic measures such as entropy, information gain, and Gini metrics to decide which optimal features should be used to separate data at each decision node. In each step, more coherent data are grouped based on the decision node. The algorithm also uses a pruning algorithm to remove unwanted sub-branches that do not contribute significantly to the decision process. In this paper, the C4.5 algorithm was not used because it uses gain ratios instead of gains, which can create more generalized trees and not fall into overfitting and it can also handle incomplete data very well (Myles et al., 2004).

NB classifier is an efficient probabilistic classifier based on Bayes' theorem of conditional probability. It is "naïve" because it assumes that all predictor variables are conditionally independent. This assumption is often violated in real-world examples, despite that NB can work well in comparison to other classifiers. The NB algorithm can work well with huge data because it needs less computation processes with reasonable speed and accuracy (Soria et al., 2011). LR is a binary classifier based on the assumption of the probabilistic statistical regression technique. LR is used to explain the relationship between one dependent binary variable and one or more independent variables by fitting examples to a logistic curve. The LR uses the sigmoid function to estimate the class probability of a test example (Kleinbaum & Klein, 2010).

Random Forest is an ensemble learning algorithm that can be constructed from a set of decision trees using the Bagging algorithm. RF adds additional randomness to the model while growing the trees. Instead of searching for the most important feature while splitting a node, it searches for the best feature among a random subset of features. This results in a wide diversity that generally results in a better model. Therefore, in a random forest, only a random subset of the features is taken into consideration by the algorithm for splitting a node. You can even make trees more random by additionally using random thresholds for each feature rather than searching for the best possible thresholds (like a normal decision tree does) (Qi, 2012).

The AdaBoost algorithm is another ensemble algorithm used to learn from multiple weak learners. The basic idea is to build a strong learner from the mistakes of several weaker learners. In other words, one could start by creating a model from the training data. Then, the second model is created from the previous one by trying to reduce the errors. Models are added sequentially, each correcting its predecessor, until the training data is predicted perfectly, or the maximum number of models have been added (Rätsch et al., 2001).

*3.4 Evaluation Measures*

Choosing the right evaluation metrics is not an easy process because the employed dataset might suffer from imbalanced data distribution which makes most predictions biased towards the dominant class label. For example, the classification accuracy metric cannot tell the true accuracy of the prediction model because it is easily influenced by a truly positive and true negative of the dominant label. Therefore, multiple evaluation metrics are highly recommended in that case. The classification evaluation metrics which were used are: Recall as shown in equation 1, Precision as shown in Eq. (2), F1 as shown in Eq. (3), Accuracy as shown in equation 4, and Area Under Curve. Recall measure computes the proportion of the observations that are correctly predicted for each class label. Precision measure computes the proportion of the observations that are tested correctly for each class label. F1 is the balance measure between precision and recall. The area under the curve measures the area under the ROC plot for each class label. Both F1 and Area Under Curve would be a good choice if data is not evenly distributed.

$$Recall = \frac{TP}{TP + FN} \quad (1)$$

$$Precision = \frac{TP}{TP + FP} \quad (2)$$

$$F1 = 2 \times \frac{(Recall \times Precision)}{Recall + Precision} \quad (3)$$

$$Accuracy = \frac{TP + TN}{TP + TN + FP + FN} \quad (4)$$

where TP (True Positive) is the number of positive observations that are predicted as such. TN (True Negative) is the number of negative observations that are predicted as such. FP (False Positive) is the number of negative observations that are predicted as positive. FN (False Negative) is the number of positive observations that are predicted as negative.

## 4. Methodology

To examine the stability of machine learning algorithms in predicting dementia groups, two feature reduction methods were used that provide flexibility in changing the feature space. These methods are Information Gain (IG) and Principal Component Analysis (PCA). Information gain attempts to rank features based on measuring the information gain metric between the output feature and each input feature then a specific threshold is applied to select top predictive features.

PCA is a feature dimensionality reduction attempt to reduce feature space into a smaller space by transforming a large set of features into a smaller one that still contains most of the information in the large set. Reducing feature space into a smaller space can reduce running costs but might affect expenses of accuracy. However, the trick is to trade a little accuracy for simplicity to make the data easier for exploration, analysis, and model construction.

Multiple experiments were conducted to investigate the stability of Dementia prediction models. For each machine learning algorithm, both IG and PCA were applied to the employed dataset by changing the threshold. For the IG method, various thresholds were applied to select top-ranked features ranging from 20% to 100%. For example, for a threshold of 20%, the top 20% of all features in the dataset were selected. For PCA, different feature space dimensions were applied to range from 2 to 8. For each threshold and dimension value, a machine-learning algorithm was applied to the reduced dataset, and evaluation measures were recorded also. Each model is validated by using 10-Fold cross-validation which separate the reduced dataset into 10 subsets. In each run, one data subset is used as testing and the remaining data subsets are used for training. Meanwhile, the evaluation measures are recorded for each model.

## 5. Results and discussion

This section presents the results obtained after conducting multiple experiments on the Dementia data set using different machine learning algorithms. The main objectives of these experiments are twofold: 1) to investigate the performance of the employed machine learning and 2) investigate their stability when changing feature space. In this study, the Information Gain was consulted in the first part of the empirical evaluation to reduce feature space by changing the threshold of selection from 20% to 100% with an increment of 10%. Initially, all features are ranked based on their Information Gain score from high to low. The process of selecting optimal features depends on assigning a threshold for the targeted score so any feature that falls within the top percentage is then selected. In other words, for 20%, the top 20% ranked features out of all features were selected. Each time a threshold was applied, and the top predictive features that are ranked by the Information Gain feature selection algorithm were selected. These features are supposed to have a great influence on detecting the group type of dementia disorder. Table 2 shows all features ranked according to the Information gain algorithm as explained in the methodology section. The threshold columns denoted by 20%, 30%, and so forth represent the features selected by that threshold – the selected features are denoted by ✓ symbol. For example, for the 20% column, it can be observed that only CDR and MMSE features have been selected and used to construct machine learning models. One might ask why the features that are not relevant to the output variable were used, the answer is that the researchers need to examine the stability of machine learning models with and without irrelevant features.

**Table 2**
Feature Ranking and selection according to Information Gain. The ✓ means that the feature has been selected by the threshold mentioned in the column header

| Rank | Feature | IG value | 20% | 30% | 40% | 50% | 60% | 70% | 80% | 90% | 100% |
|---|---|---|---|---|---|---|---|---|---|---|---|
| 1 | CDR | 0.724 | ✓ | ✓ | ✓ | ✓ | ✓ | ✓ | ✓ | ✓ | ✓ |
| 2 | MMSE | 0.336 | ✓ | ✓ | ✓ | ✓ | ✓ | ✓ | ✓ | ✓ | ✓ |
| 3 | M/F | 0.070 | | ✓ | ✓ | ✓ | ✓ | ✓ | ✓ | ✓ | ✓ |
| 4 | nWBV | 0.045 | | | ✓ | ✓ | ✓ | ✓ | ✓ | ✓ | ✓ |
| 5 | EDUC | 0.036 | | | | ✓ | ✓ | ✓ | ✓ | ✓ | ✓ |
| 6 | SES | 0.030 | | | | ✓ | ✓ | ✓ | ✓ | ✓ | ✓ |
| 7 | MR Delay | 0.023 | | | | | ✓ | ✓ | ✓ | ✓ | ✓ |
| 8 | Visit | 0.008 | | | | | | ✓ | ✓ | ✓ | ✓ |
| 9 | ASF | 0.003 | | | | | | | ✓ | ✓ | ✓ |
| 10 | eTIV | 0.002 | | | | | | | | ✓ | ✓ |
| 11 | Age | 0.002 | | | | | | | | | ✓ |

Tables 3 to 7 show the aggregate evaluation results for all labels using different selection thresholds of top predictive features. The values in the bold represent the most predictive models for each feature set that are selected by a threshold. Table 3 shows the results of all machine learning models in terms of the AUC measure. It can be observed that no one model can perform better than other models. Remarkably, both DT and NB can show somehow more accurate results than other models. This has also been confirmed by the results in Figure 5 that shows values of AUC for each machine learning model across different threshold values. DT and NB show stable accuracy which suggests that the changes in AUC across different thresholds. Although SVM can use different kernel functions to map data from low dimension to higher, it does not work efficiently for this complex structure dataset because it contains numerical and categorical data. All used models are supposed to support the nonlinear relationship between variables, but since SVM cannot treat input categorical variables in its implementation, only numeric input variables were used. This may explain why SVM could not surpass other prediction models. Surprisingly, the RF obtained less accurate results, even though that RF can treat both numerical and categorical variables and built upon the idea of ensemble learning.

**Table 3**
Results of AUC evaluation measure against different feature sets

| Model | %of selected features by Information Gain | | | | | | | | |
|---|---|---|---|---|---|---|---|---|---|
| | 20% | 30% | 40% | 50% | 60% | 70% | 80% | 90% | 100% |
| RF | 0.919 | 0.955 | 0.896 | 0.930 | 0.901 | 0.890 | 0.903 | 0.893 | 0.901 |
| LR | 0.927 | 0.955 | 0.917 | 0.930 | 0.890 | 0.850 | 0.852 | 0.842 | 0.885 |
| NB | 0.940 | 0.953 | 0.953 | 0.954 | 0.963 | 0.963 | 0.966 | 0.967 | 0.967 |
| SVM | 0.935 | 0.933 | 0.956 | 0.956 | 0.959 | 0.960 | 0.960 | 0.961 | 0.960 |
| DT | 0.924 | 0.925 | 0.947 | 0.958 | 0.964 | 0.963 | 0.967 | 0.964 | 0.967 |
| kNN | 0.938 | 0.903 | 0.931 | 0.951 | 0.956 | 0.956 | 0.960 | 0.958 | 0.958 |
| Ada | 0.927 | 0.875 | 0.933 | 0.926 | 0.941 | 0.940 | 0.909 | 0.906 | 0.931 |

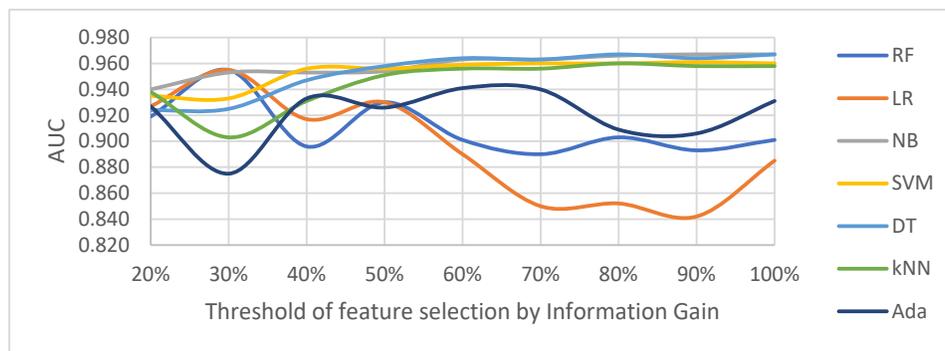

**Fig. 5.** AUC results for the comparison between different machine learning models using IG

According to Fig. 5, it can be noticed that LR, RF, and Ada are the most unstable models, whereas NB and DT are the most stable models with slight changes in AUC while changing the selection threshold. Also, there is an improvement in AUC when selecting more features despite some selected features are irrelevant according to Information Gain. Finally, the kNN model shows less accurate AUC results when dealing with a small threshold value, but the AUC improves when the threshold value increases. To investigate more on why kNN behaves badly over small threshold values, the researchers have tried to change the number of $k$ nearest neighbors from 1 to 10 but still, the best result for kNN is so far from other results for small threshold values. Another factor that may increase the accuracy of kNN is the choice of feature weights during the voting process. Two approaches were tried; the first approach is using equal weights for all retrieved $k$ nearest neighbors and the second approach is to use their distance to increase the weight of some labels over others. The latter gives us better results. Indeed, searching for optimal configurations within a large space of configuration possibilities is a time-consuming task and needs optimization algorithms. However, the recorded results for kNN are those after searching manually for the best configurations that include *k* and weights. Surprisingly, the RF and AdaBoost models that are considered as ensemble algorithms did not beat DT across different feature selection thresholds. Tables 4 to 7 show results for other evaluation measures such as Accuracy, F1, Precision, and Recall. The results in these tables are relatively similar which suggests that kNN, DT, and NB are the most stable models with high values. These results are also confirmed by findings in Figures 6 to 9. In conclusion, it can be found that ensemble methods such as RF and Ada boost cannot beat solo models such as NB, DT, and kNN. Also, the LR model obtained the worst results among different models. The solo models show improvements while increasing the selection threshold which confirms that using as many features as possible increases accuracy in spite that some of these features are irrelevant.

**Table 4**
Results of Accuracy evaluation measure against different feature sets

| Model | %of selected features by Information Gain | | | | | | | | |
|---|---|---|---|---|---|---|---|---|---|
| | 20% | 30% | 40% | 50% | 60% | 70% | 80% | 90% | 100% |
| RF | 0.944 | 0.928 | 0.871 | 0.930 | 0.901 | 0.890 | 0.903 | 0.893 | 0.901 |
| LR | 0.928 | 0.946 | 0.917 | 0.930 | 0.890 | 0.850 | 0.853 | 0.842 | 0.885 |
| NB | 0.946 | 0.946 | 0.946 | 0.946 | 0.941 | 0.938 | 0.938 | 0.938 | 0.938 |
| SVM | 0.946 | 0.946 | 0.946 | 0.941 | 0.936 | 0.936 | 0.938 | 0.941 | 0.941 |
| DT | 0.946 | 0.917 | 0.933 | 0.936 | 0.944 | 0.946 | 0.946 | 0.946 | 0.938 |
| kNN | 0.946 | 0.903 | 0.946 | 0.946 | 0.946 | 0.946 | 0.946 | 0.946 | 0.946 |
| Ada | 0.946 | 0.815 | 0.917 | 0.906 | 0.903 | 0.903 | 0.901 | 0.898 | 0.903 |

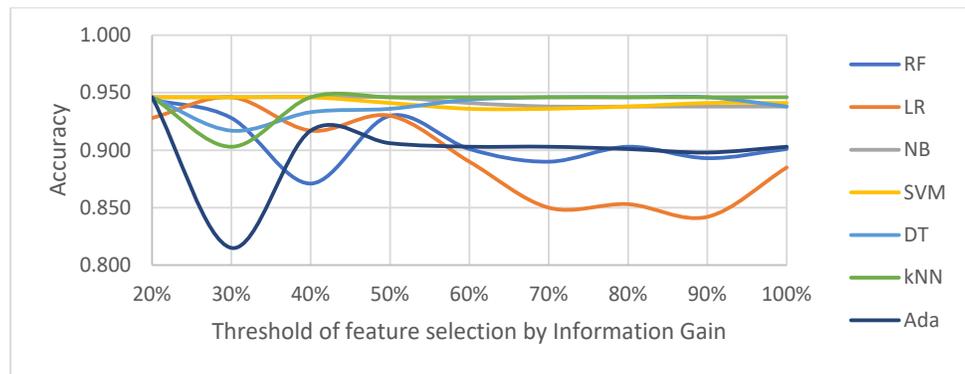

**Fig. 6.** Accuracy results for the comparison between different machine learning models using IG

**Table 5**
Results of F1 evaluation measure against different feature sets

| Model | %of selected features by Information Gain | | | | | | | | |
|---|---|---|---|---|---|---|---|---|---|
| | 20% | 30% | 40% | 50% | 60% | 70% | 80% | 90% | 100% |
| RF | 0.944 | 0.928 | 0.871 | 0.930 | 0.901 | 0.890 | 0.903 | 0.893 | 0.901 |
| LR | 0.928 | 0.946 | 0.917 | 0.930 | 0.890 | 0.850 | 0.852 | 0.842 | 0.885 |
| NB | 0.946 | 0.946 | 0.946 | 0.946 | 0.941 | 0.938 | 0.938 | 0.938 | 0.938 |
| SVM | 0.946 | 0.946 | 0.946 | 0.941 | 0.936 | 0.936 | 0.938 | 0.941 | 0.941 |
| DT | 0.946 | 0.917 | 0.933 | 0.936 | 0.944 | 0.946 | 0.946 | 0.946 | 0.938 |
| kNN | 0.946 | 0.903 | 0.946 | 0.946 | 0.946 | 0.946 | 0.946 | 0.946 | 0.946 |
| Ada | 0.946 | 0.815 | 0.917 | 0.906 | 0.903 | 0.903 | 0.901 | 0.898 | 0.903 |

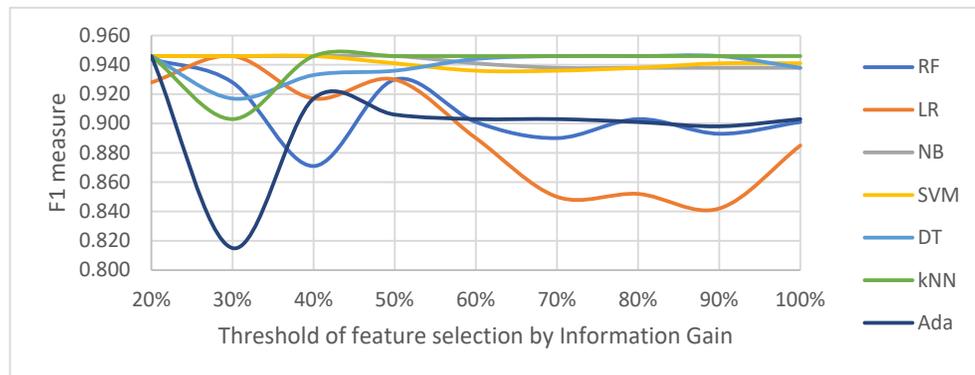

**Fig. 7.** F1 results for the comparison between different machine learning models using IG

**Table 6**
Results of Precision evaluation measure against different feature sets

| Model | %of selected features by Information Gain | | | | | | | | |
|---|---|---|---|---|---|---|---|---|---|
| | 20% | 30% | 40% | 50% | 60% | 70% | 80% | 90% | 100% |
| RF | 0.944 | 0.928 | 0.871 | 0.930 | 0.901 | 0.890 | 0.903 | 0.893 | 0.901 |
| LR | 0.928 | 0.946 | 0.917 | 0.930 | 0.890 | 0.850 | 0.853 | 0.842 | 0.885 |
| NB | 0.946 | 0.946 | 0.946 | 0.946 | 0.941 | 0.938 | 0.938 | 0.938 | 0.938 |
| SVM | 0.946 | 0.946 | 0.946 | 0.941 | 0.936 | 0.936 | 0.938 | 0.941 | 0.941 |
| DT | 0.946 | 0.917 | 0.933 | 0.936 | 0.944 | 0.946 | 0.946 | 0.946 | 0.938 |
| kNN | 0.946 | 0.903 | 0.946 | 0.946 | 0.946 | 0.946 | 0.946 | 0.946 | 0.946 |
| Ada | 0.946 | 0.815 | 0.917 | 0.906 | 0.903 | 0.903 | 0.901 | 0.898 | 0.903 |

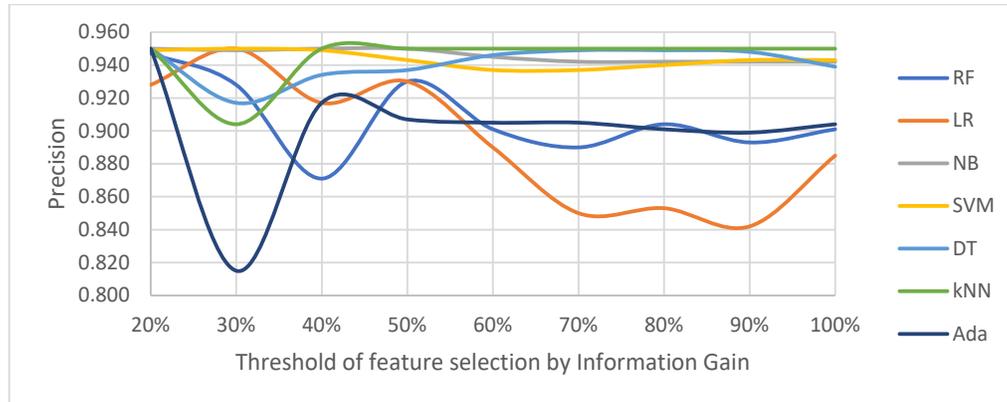

**Fig. 8.** Precision results for the comparison between different machine learning models using IG

In this study, the second part of the empirical evaluation is to further examine the stability of the constructed prediction models in accurately predicting the Dementia group by using the PCA algorithm. The purpose of this validation is to ensure whether the accuracy of prediction models improve and keep stable or there are some fluctuations due to changes in the dimension of generated feature space. Specifically, The PCA algorithm is used to generate a new feature space with a different dimension that changes from 2 to 8. For each dimension, the empirical experiments were re-run on the new dataset and the same machine learning algorithms. The obtained results are recorded in Tables 8 to 12. The first remarkable observation is that all evaluation results are bad when the dimension of feature space is small, and these results are improved along with an increasing dimension of feature space. Furthermore, the PCA results, in general, are worse than the results of the first experiments that use Information gain for feature selection. This is because the PCA generates a new feature space that does not exist in the original dataset.

**Table 7**
Results of Recall evaluation measure against different feature sets

| Model | %of selected features by Information Gain | | | | | | | | |
|---|---|---|---|---|---|---|---|---|---|
| | 20% | 30% | 40% | 50% | 60% | 70% | 80% | 90% | 100% |
| RF | 0.947 | 0.928 | 0.871 | 0.930 | 0.901 | 0.890 | 0.904 | 0.893 | 0.901 |
| LR | 0.928 | 0.950 | 0.917 | 0.930 | 0.890 | 0.850 | 0.853 | 0.842 | 0.885 |
| NB | 0.950 | 0.949 | 0.950 | 0.950 | 0.945 | 0.942 | 0.942 | 0.942 | 0.942 |
| SVM | 0.949 | 0.950 | 0.949 | 0.943 | 0.937 | 0.937 | 0.940 | 0.943 | 0.943 |
| DT | 0.950 | 0.917 | 0.934 | 0.937 | 0.946 | 0.949 | 0.949 | 0.948 | 0.939 |
| kNN | 0.950 | 0.904 | 0.950 | 0.950 | 0.950 | 0.950 | 0.950 | 0.950 | 0.950 |
| Ada | 0.950 | 0.815 | 0.917 | 0.907 | 0.905 | 0.905 | 0.901 | 0.899 | 0.904 |

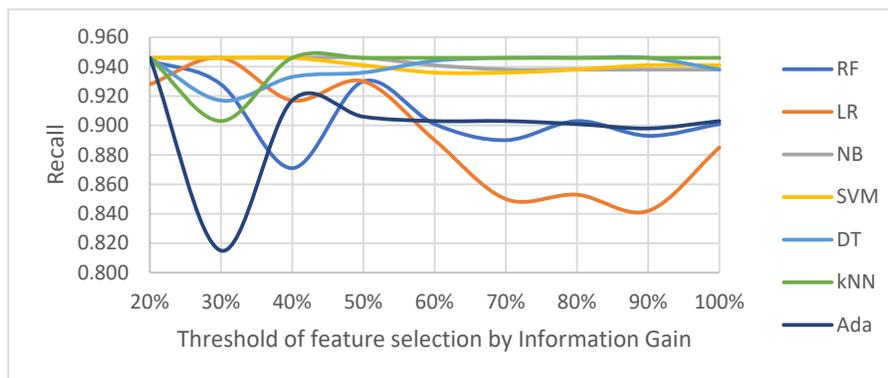

**Fig. 9.** Recall results for the comparison between different machine learning models using IG

342For AUC in Table 8, it can be observed that NB obtained accurate and stable results across different feature space dimension values. For the Accuracy metric in Table 9, it can be seen that NB and kNN are more predictive models. Similarly, Tables 10, 11, and 12 show the same trend exists in Table 9.

**Table 8**
Results of AUC evaluation measure against different feature dimensions generated by PCA

| Model | Feature dimension generated by PCA | | | | | | |
|---|---|---|---|---|---|---|---|
| | 2 | 3 | 4 | 5 | 6 | 7 | 8 |
| RF | 0.557 | 0.543 | 0.708 | 0.766 | 0.817 | 0.825 | 0.822 |
| LR | 0.565 | 0.600 | 0.710 | 0.775 | 0.841 | 0.863 | 0.863 |
| NB | 0.656 | 0.682 | 0.835 | 0.931 | 0.962 | 0.967 | 0.967 |
| SVM | 0.630 | 0.664 | 0.805 | 0.888 | 0.905 | 0.917 | 0.917 |
| DT | 0.633 | 0.623 | 0.783 | 0.901 | 0.934 | 0.942 | 0.938 |
| kNN | 0.554 | 0.608 | 0.803 | 0.927 | 0.955 | 0.961 | 0.964 |
| Ada | 0.583 | 0.585 | 0.677 | 0.759 | 0.819 | 0.810 | 0.825 |

**Table 9**
Results of Accuracy evaluation measure against different feature dimensions generated by PCA

| Model | Feature dimension generated by PCA | | | | | | |
|---|---|---|---|---|---|---|---|
| | 2 | 3 | 4 | 5 | 6 | 7 | 8 |
| RF | 0.558 | 0.544 | 0.708 | 0.767 | 0.818 | 0.826 | 0.823 |
| LR | 0.566 | 0.601 | 0.710 | 0.775 | 0.842 | 0.863 | 0.863 |
| NB | 0.625 | 0.622 | 0.764 | 0.847 | 0.925 | 0.933 | 0.933 |
| SVM | 0.603 | 0.609 | 0.737 | 0.815 | 0.845 | 0.855 | 0.850 |
| DT | 0.622 | 0.582 | 0.708 | 0.828 | 0.898 | 0.874 | 0.863 |
| kNN | 0.576 | 0.563 | 0.751 | 0.861 | 0.949 | 0.946 | 0.946 |
| Ada | 0.579 | 0.576 | 0.670 | 0.745 | 0.812 | 0.810 | 0.818 |

**Table 10**
Results of F1 evaluation measure against different feature dimensions generated by PCA

| Model | Feature dimension generated by PCA | | | | | | |
|---|---|---|---|---|---|---|---|
| | 2 | 3 | 4 | 5 | 6 | 7 | 8 |
| RF | 0.557 | 0.543 | 0.708 | 0.767 | 0.818 | 0.826 | 0.823 |
| LR | 0.565 | 0.600 | 0.710 | 0.775 | 0.842 | 0.863 | 0.863 |
| NB | 0.622 | 0.620 | 0.764 | 0.847 | 0.925 | 0.933 | 0.933 |
| SVM | 0.600 | 0.605 | 0.737 | 0.814 | 0.844 | 0.854 | 0.849 |
| DT | 0.621 | 0.582 | 0.707 | 0.828 | 0.898 | 0.874 | 0.863 |
| kNN | 0.573 | 0.563 | 0.751 | 0.860 | 0.949 | 0.946 | 0.946 |
| Ada | 0.577 | 0.576 | 0.670 | 0.745 | 0.812 | 0.810 | 0.818 |

**Table 11**
Results of Recall evaluation measure against different feature dimensions generated by PCA

| Model | Feature dimension generated by PCA | | | | | | |
|---|---|---|---|---|---|---|---|
| | 2 | 3 | 4 | 5 | 6 | 7 | 8 |
| RF | 0.558 | 0.544 | 0.708 | 0.767 | 0.818 | 0.826 | 0.823 |
| LR | 0.566 | 0.601 | 0.710 | 0.775 | 0.842 | 0.863 | 0.863 |
| NB | 0.625 | 0.622 | 0.764 | 0.847 | 0.925 | 0.933 | 0.933 |
| SVM | 0.603 | 0.609 | 0.737 | 0.815 | 0.845 | 0.855 | 0.850 |
| DT | 0.622 | 0.582 | 0.708 | 0.828 | 0.898 | 0.874 | 0.863 |
| kNN | 0.576 | 0.563 | 0.751 | 0.861 | 0.949 | 0.946 | 0.946 |
| Ada | 0.579 | 0.576 | 0.670 | 0.745 | 0.812 | 0.810 | 0.818 |

**Table 12**
Results of Precision evaluation measure against different feature dimension generated by PCA

| Model | Feature dimension generated by PCA | | | | | | |
|---|---|---|---|---|---|---|---|
| | 2 | 3 | 4 | 5 | 6 | 7 | 8 |
| RF | 0.557 | 0.544 | 0.708 | 0.767 | 0.818 | 0.826 | 0.824 |
| LR | 0.565 | 0.600 | 0.711 | 0.775 | 0.843 | 0.863 | 0.863 |
| NB | 0.626 | 0.623 | 0.764 | 0.849 | 0.928 | 0.937 | 0.937 |
| SVM | 0.604 | 0.611 | 0.737 | 0.817 | 0.850 | 0.861 | 0.854 |
| DT | 0.622 | 0.582 | 0.708 | 0.830 | 0.902 | 0.876 | 0.863 |
| kNN | 0.577 | 0.564 | 0.751 | 0.863 | 0.952 | 0.950 | 0.950 |
| Ada | 0.583 | 0.577 | 0.670 | 0.746 | 0.812 | 0.810 | 0.818 |

Figs. (10-14) show the relationship between evaluation measure and feature space dimension for each machine learning model. The findings in these figures suggest that all models behave similarly with a bad performance at small feature space dimensions and these results getting improved while the dimension is increased.

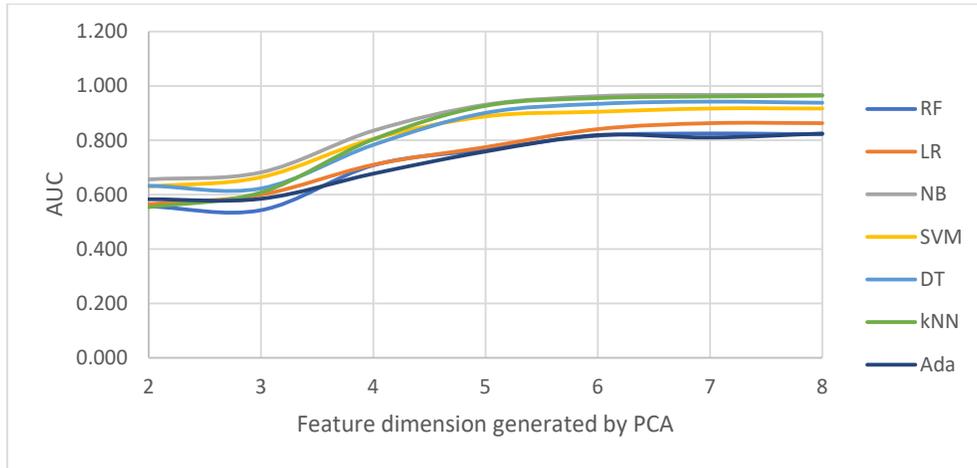

**Fig. 10.** AUC results for the comparison between different machine learning models using PCA

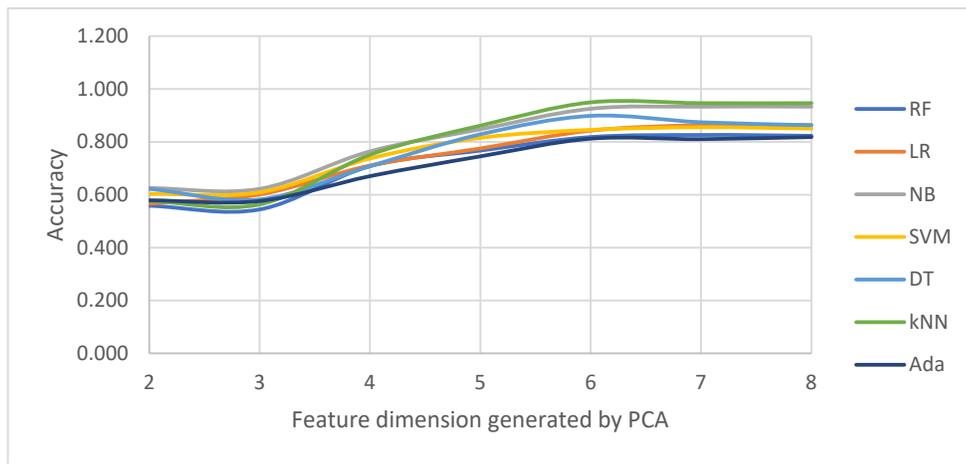

**Fig. 11.** Accuracy results for the comparison between different machine learning models using PCA

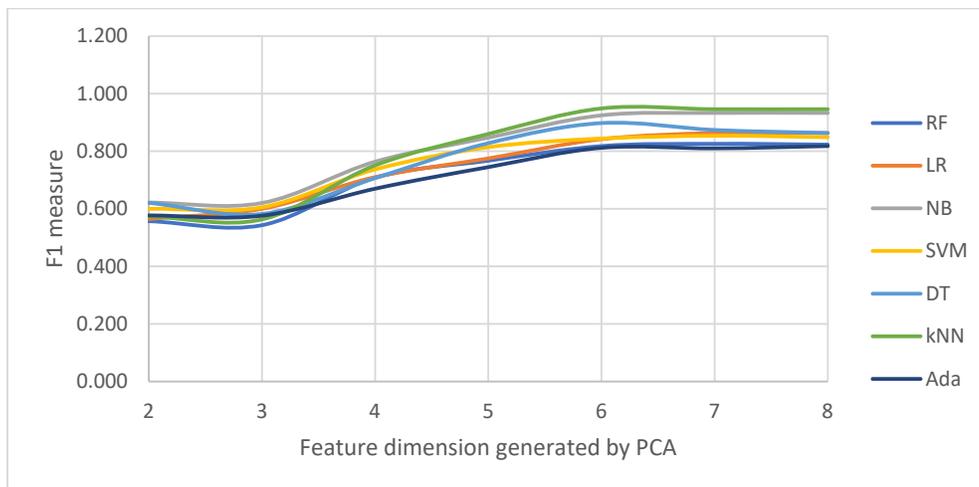

**Fig. 12.** F1 results for the comparison between different machine learning models using PCA

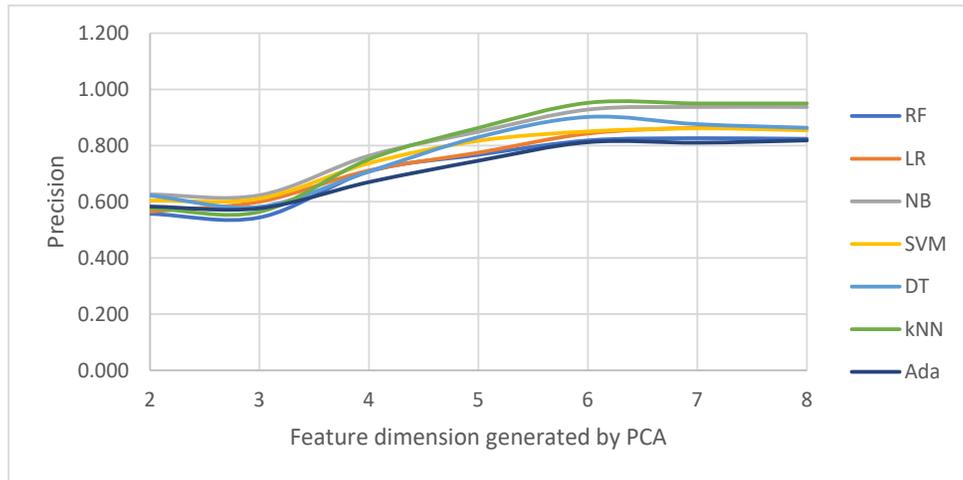

**Fig. 13.** Precision results for the comparison between different machine learning models using PCA

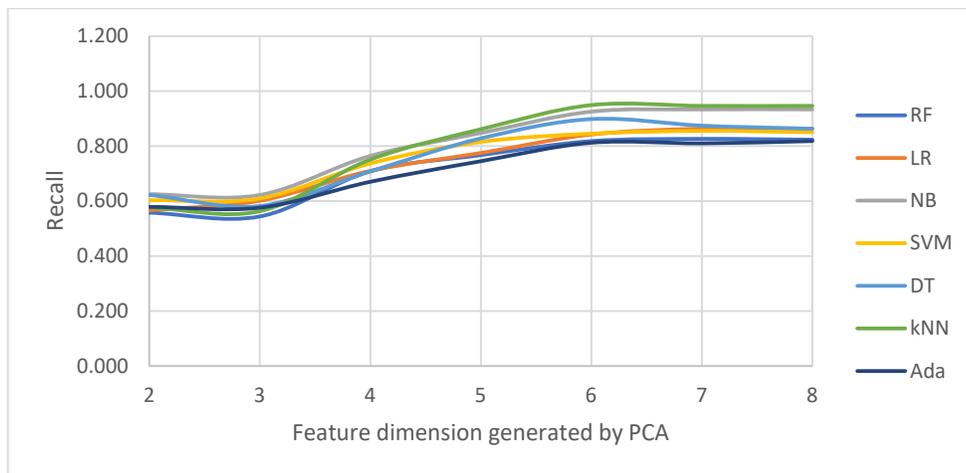

**Fig. 14.** Recall results for the comparison between different machine learning models using PCA

## 6. Conclusion

This paper investigates the stability of machine learning methods for predicting dementia in older people. Seven machine learning algorithms in addition to two feature reduction methods with different selection thresholds have been applied to the Dementia dataset. In each experiment, a new set of features were applied as explained in the methodology section. The results drawn from this comprehensive experimentation are: 1) Using Information gain to rank and select features would seem to be more efficient and stable than using PCA. 2) In general, there is no clear stability of all machine learning methods concerning all accuracy measures. 3) It was noticed that Support Vector Machine and Naïve Bayes are the top stable methods among all used machine learning methods. 4) Using high feature dimensions in PCA is better than using a small number of dimensions. 5) Surprisingly, ensemble learning methods such as Ada and Random forest did not show stable performance under different accuracy measures. 6) Finally, adding more features by incrementing the threshold of information gain resulted in an improvement in solo models although some of the added features may be irrelevant. The algorithms that showed consistent improvements are Support Vector Machine, Naïve Bayes, Decision tree, and k-Nearest Neighbors, while the ensemble methods such as Ada and Random Forest did not improve. From these results, it can be concluded that the Support Vector Machine and Naïve Bayes are the most stable methods. However, we doubt that such methods did not work well with low feature dimensionality even though all their accuracy values are superior to other methods. Also, we discourage using ensemble learning methods for predicting Dementia, under any conditional setting.

**Acknowledgment**

The authors are grateful to the Applied Science Private University, Amman-Jordan, for the full financial support granted to cover the publication fee of this research article.